\documentclass[runningheads]{llncs}

\usepackage{eccv}

\usepackage{eccvabbrv}
\usepackage{makecell}
\usepackage{graphicx}
\usepackage{multirow}
\usepackage{comment}
\usepackage{wrapfig}
\usepackage{lipsum}

\usepackage[accsupp]{axessibility}  

\usepackage{booktabs}
\usepackage{amssymb}
\usepackage{threeparttable} 

\usepackage[breaklinks,colorlinks,citecolor=eccvblue]{hyperref}
\usepackage{xspace}

\usepackage{orcidlink}
\usepackage{CJK}

\newcommand{\contributionfont}
{\fontsize{9}{9}\selectfont}

\newcommand{\equalmark}{\textsuperscript{$*$}}
\newcommand{\leadmark}{\textsuperscript{$\ddagger$}}
\newcommand{\correspondmark}{\textsuperscript{$\dagger$}}

\newcommand{\benchname}{GRADE\xspace}

\begin{document}


\title{\benchname: Benchmarking Discipline-Informed Reasoning in Image Editing}

\titlerunning{\benchname}

\author{Mingxin Liu\inst{1}\equalmark\orcidlink{0000-1111-2222-3333} \and
Ziqian Fan\inst{2}\equalmark\orcidlink{1111-2222-3333-4444}
\and
Zhaokai Wang\inst{1}\equalmark\leadmark\orcidlink{2222--3333-4444-5555}
\and
Leyao Gu\inst{1}\equalmark\orcidlink{2222--3333-4444-5555}
\and
Zirun Zhu\inst{1}\equalmark\orcidlink{2222--3333-4444-5555}
\and
Yiguo He\inst{1}\orcidlink{2222--3333-4444-5555}
\and
Yuchen Yang\inst{3}\orcidlink{2222--3333-4444-5555}
\and
Changyao Tian\inst{4}\orcidlink{2222--3333-4444-5555}
\and
Xiangyu Zhao\inst{1}\orcidlink{2222--3333-4444-5555}
\and
Ning Liao\inst{1}\orcidlink{2222--3333-4444-5555}
\and
Shaofeng Zhang\inst{5}\orcidlink{2222--3333-4444-5555}
\and
Qibing Ren\inst{1}\orcidlink{2222--3333-4444-5555}
\and
Zhihang Zhong\inst{1}\orcidlink{2222--3333-4444-5555}
\and
Xuanhe Zhou\inst{1}\orcidlink{2222--3333-4444-5555}
\and
Junchi Yan\inst{1}\orcidlink{2222--3333-4444-5555}
\and
Xue Yang\inst{1}\correspondmark\orcidlink{2222--3333-4444-5555}
}

\authorrunning{M.~Liu et al.}

\institute{Shanghai Jiao Tong University \and
South China University of Technology 
\and
Fudan University
\and
The Chinese University of Hong Kong
\and
University of Science and Technology of China}
\maketitle
\vspace{-3mm}
\begin{center}
\noindent{\contributionfont
\equalmark Core contributors \quad
\leadmark Project lead \quad
\correspondmark Corresponding author
}   
\vspace{3mm}

Project Page: \url{https://grade-bench.github.io/}

Evaluation Code: \url{https://github.com/VisionXLab/GRADE}

Benchmark: \url{https://huggingface.co/datasets/VisionXLab/GRADE}

\end{center}











\begin{abstract}
Unified multimodal models target joint understanding, reasoning, and generation, but current image editing benchmarks are largely confined to natural images and shallow commonsense reasoning, offering limited assessment of this capability under structured, domain-specific constraints.
In this work, we introduce GRADE, the first benchmark to assess discipline-informed knowledge and reasoning in image editing. 
GRADE comprises 520 carefully curated samples across 10 academic domains, spanning from natural science to social science.
To support rigorous evaluation, we propose a multi-dimensional evaluation protocol that jointly assesses Discipline Reasoning, Visual Consistency, and Logical Readability. 
Extensive experiments on 20 state-of-the-art open-source and closed-source models reveal substantial limitations in current models under implicit, knowledge-intensive editing settings, leading to large performance gaps.
Beyond quantitative scores, we conduct rigorous analyses and ablations to expose model shortcomings and identify the constraints within disciplinary editing.
Together, GRADE pinpoints key directions for the future development of unified multimodal models, advancing the research on discipline-informed image editing and reasoning. Our benchmark and evaluation code are publicly released.




\end{abstract}

\section{Introduction}
\label{sec:intro}

\begin{figure*}[!tb]

    \centering
    
    \includegraphics[width=\linewidth]{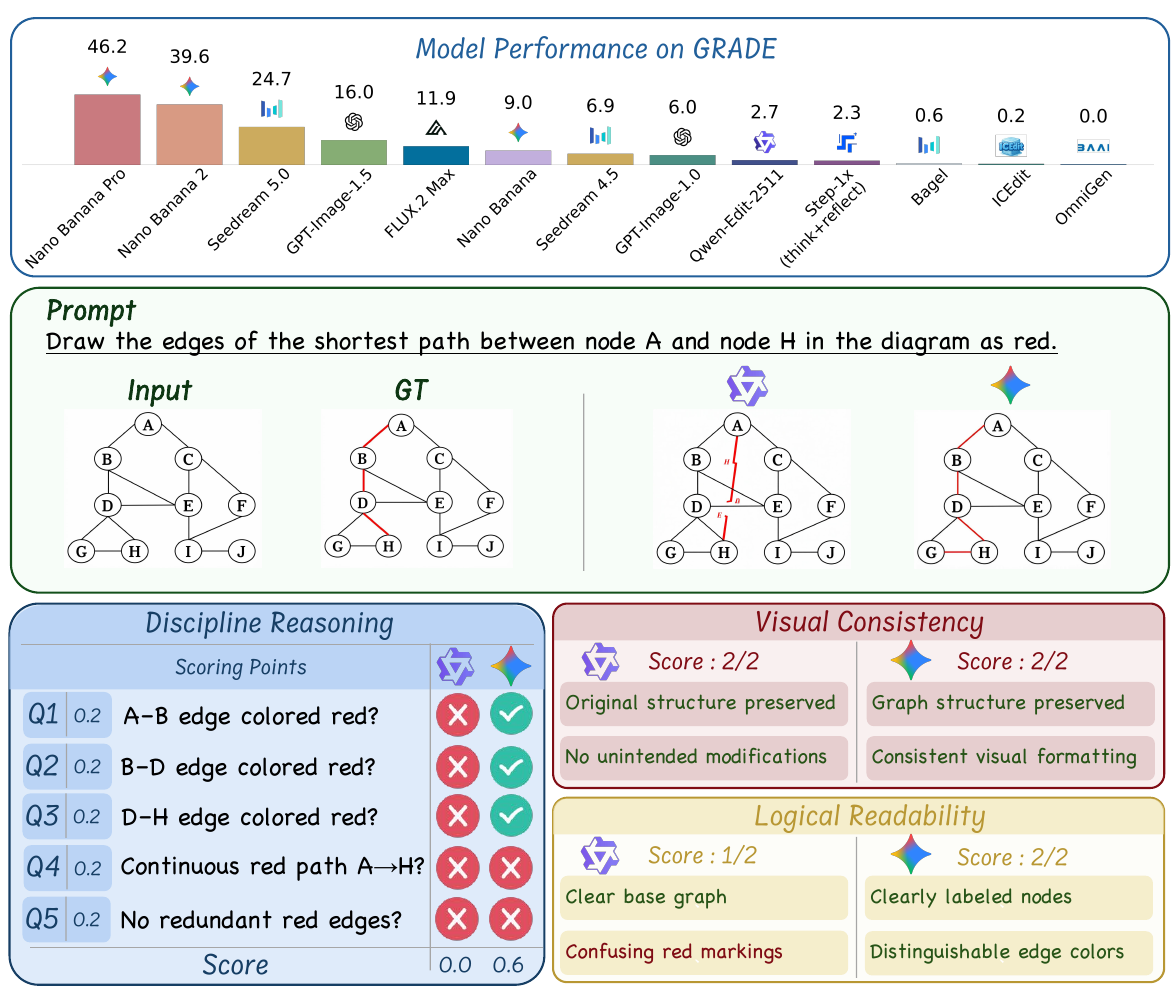}
  
    \caption{\textbf{Examples of images generated by state-of-the-art editing models on \benchname and their evaluation results.} Challenging discipline-informed image editing exposes limitations of current models in complex knowledge reasoning. Notable performance gaps exist across models on GRADE. 
    }
    \label{fig:teaser}
\end{figure*}

Recent years have witnessed rapid advances in unified multimodal models (UMMs), \textit{i.e.} models with unified abilities of multimodal understanding and image generation~\cite{team2024chameleon,deng2025emerging,GPT-Image-1.5,Nano-Banana-Pro}.
Modern UMMs are expected to integrate knowledge, structured reasoning, and controllable generation within a single system. In this context, several existing benchmarks~\cite{zhao2025envisioning, sun2025t2ireasonbenchbenchmarkingreasoninginformedtexttoimage} attempt to evaluate reasoning in image editing. They are predominantly grounded in natural image domains, where reasoning difficulty mainly arises from the implicitness or linguistic complexity of prompts, and the underlying knowledge involved is typically shallow, relying largely on everyday commonsense. As a result, these settings provide only a limited test of whether unified multimodal models can truly coordinate knowledge, reasoning, and editing in a principled manner. 

Inspired by prior work that introduces discipline-specific knowledge as a challenging axis in multimodal understanding~\cite{yue2024mmmu,phan2025humanity} and text-to-image  generation~\cite{luo2025mmmg,wang2025genexam} tasks, we argue that discipline-informed image editing offers a more stringent and representative evaluation setting, as it requires models to operate over structured discipline-informed knowledge, which is inherently more demanding than everyday commonsense. In contrast to prior evaluation settings that assess discipline-specific knowledge grounding either through pure image understanding or unconditioned image generation, image editing requires models to reason under domain constraints while preserving existing visual structures and making precise modifications, suggesting greater challenges for UMMs.


Motivated by this gap, we introduce GRADE, which stands for \textbf{G}rounded \textbf{R}easoning \textbf{A}ssessment for
\textbf{D}iscipline-informed \textbf{E}diting. GRADE is the first image editing benchmark explicitly designed to evaluate discipline-informed reasoning. While prior benchmarks mainly emphasize the implicitness or linguistic complexity of prompts, GRADE enriches knowledge and reasoning difficulty through the challenging discipline-specific image domain.
GRADE comprises 520 carefully curated samples spanning 10 academic domains, from natural sciences to humanities and applied fields. 
They reflect practical image editing scenarios 
such as assisting researchers or teaching assistants in correcting geometric diagrams, modifying chemical structures, or refining data visualizations, thereby providing a comprehensive and challenging testbed for evaluating discipline-informed reasoning in image editing.
For the evaluation metrics, as shown in Fig.~\ref{fig:teaser}, we introduce Discipline Reasoning for explicitly measuring discipline-informed knowledge reasoning with fine-grained scoring points, Visual Consistency for task-dependent consistency constraints, and Logical Readability for evaluating logically clear and well-structured academic representations. These dimensions provide a comprehensive evaluation framework that goes beyond aesthetic quality and realism in general image editing to assess the rigor and interpretability of discipline-informed image editing.  



We evaluate 20 state-of-the-art (SoTA) image editing models on GRADE. We observe that models with comparable performance on other benchmarks, such as Nano Banana Pro~\cite{Nano-Banana-Pro} and GPT-Image-1.5~\cite{GPT-Image-1.5}, exhibit markedly different behaviors under discipline-informed editing (46.2\% vs. 16.0\%), highlighting GRADE's stronger discriminative power in probing implicit academic knowledge and structured reasoning. We also observe a huge gap between open-source and closed-source models, where the SoTA open-source model Qwen-Edit-2511~\cite{wu2025qwenimagetechnicalreport} (2.7\%) significantly lags behind closed-source SoTA. 

Overall, implicit discipline-informed reasoning remains a major bottleneck in current models. 
To further examine this limitation, we analyze representative failure cases and compare implicit and explicit instruction formulations, providing additional diagnostic insights into current model behavior. In sum, GRADE provides clear and actionable directions for the future advancement of UMMs towards discipline-informed image editing and reasoning tasks.

Overall, our contributions are as follows:

1. We present the first discipline-informed image editing benchmark spanning multiple academic domains, designed to rigorously evaluate knowledge grounding and reasoning in image editing models.

2. We propose a multi-dimensional automated evaluation pipeline that is scalable and exhibits strong alignment with human judgments.

3. We conduct comprehensive analysis across a wide range of SoTA models, revealing key limitations of current approaches and offering actionable guidance for future research.




\section{Related Work}
\subsection{Image Generation Models}
Recent advances in image generation are primarily driven by diffusion-based models~\cite{dhariwal2021diffusion,ho2020denoising}, which serve as the foundation for modern text-conditioned synthesis.
Early text-to-image systems, such as VQGAN-based pipelines and cascaded diffusion models~\cite{crowson2022vqgan, balaji2022ediff}, established the feasibility of aligning visual synthesis with natural language descriptions, while subsequent works improved photorealism and semantic faithfulness through large-scale training and refined conditioning mechanisms~\cite{saharia2022photorealistic, zhang2023adding}.

More recently, a line of research has shifted toward unified generative frameworks that aim to integrate language understanding, visual perception, and image generation within a single model~\cite{team2024chameleon,deng2025emerging,zhang2025unimodel,li2025synergen}. 
Open-source efforts such as OmniGen~\cite{xiao2025omnigen} emphasize accessibility and extensibility, enabling community-driven exploration of unified generation capabilities. 
Meanwhile, large-scale proprietary systems, including Gemini~\cite{team2023gemini}, Seedream~\cite{seedream2025seedream}, and recent OpenAI models~\cite{singh2025openai}, demonstrate strong visual quality and robustness through extensive data and infrastructure support.

In addition, image editing models~\cite{li2024surveymultimodalcompositeediting} such as Step-1X~\cite{liu2025step1x}, Qwen-Image~\cite{wu2025qwenimagetechnicalreport}, and others~\cite{zhang2025icedit} further extend text-to-image frameworks by incorporating an input image and a textual instruction to perform targeted modifications. While recent models often produce visually plausible edits, their ability to handle reasoning-intensive scenarios, particularly those requiring structured, domain-specific knowledge, remains largely unexplored.



\subsection{Image Editing Benchmarks}
To enable systematic evaluation of generative models, a range of benchmarks~\cite{ye2025imgedit,huang2023smarteditexploringcomplexinstructionbased} have been proposed to assess visual quality~\cite{heusel2018ganstrainedtimescaleupdate,salimans2016improvedtechniquestraininggans,yu2022scalingautoregressivemodelscontentrich} and semantic alignment~\cite{hu2024ellaequipdiffusionmodels,ghosh2023genevalobjectfocusedframeworkevaluating}.
However, most existing image editing benchmarks rely on highly explicit instructions that directly specify the required operations. ImgEdit~\cite{ye2025imgedit} primarily targets traditional editing tasks where the required operations are explicitly specified and reasoning is not a central concern. As UMMs rapidly advance, the community has placed increasing emphasis on evaluating their reasoning capabilities. More recent editing benchmarks begin to incorporate implicit reasoning, but the involved knowledge is typically restricted to general-purpose commonsense rather than disciplinary expertise. RISEBench~\cite{zhao2025envisioning} categorizes reasoning into temporal, causal, spatial, and logical types, while KRISBench~\cite{wu2025kris} organizes reasoning knowledge based on cognitively motivated taxonomies. While these benchmarks effectively assess general reasoning, they do not focus on the assessment of disciplinary knowledge in editing.

\subsection{Discipline-Specific Benchmarks}
Disciplinary knowledge is widely regarded as one of the most challenging forms of reasoning and has been extensively explored in previous benchmarks. In multimodal understanding, MMMU~\cite{yue2024mmmu} evaluates multimodal reasoning across six broad discipline categories spanning over thirty disciplines. Similarly, HLE~\cite{phan2025humanity} includes image-related tasks that require PhD-level, multi-disciplinary understanding, emphasizing high-difficulty academic reasoning in visual comprehension.
In the text-to-image setting, MMMG~\cite{luo2025mmmg} and Sridbench~\cite{chang2025sridbench} focus on disciplinary concept illustration, and GenExam~\cite{wang2025genexam} further explores generation tasks grounded in professional disciplinary knowledge. Despite these advances, disciplinary reasoning remains largely unexplored in the image editing setting, which requires tighter integration of knowledge understanding, reasoning, and editing capabilities.

\begin{figure*}[tb!]

    \centering
    \includegraphics[width=\linewidth]{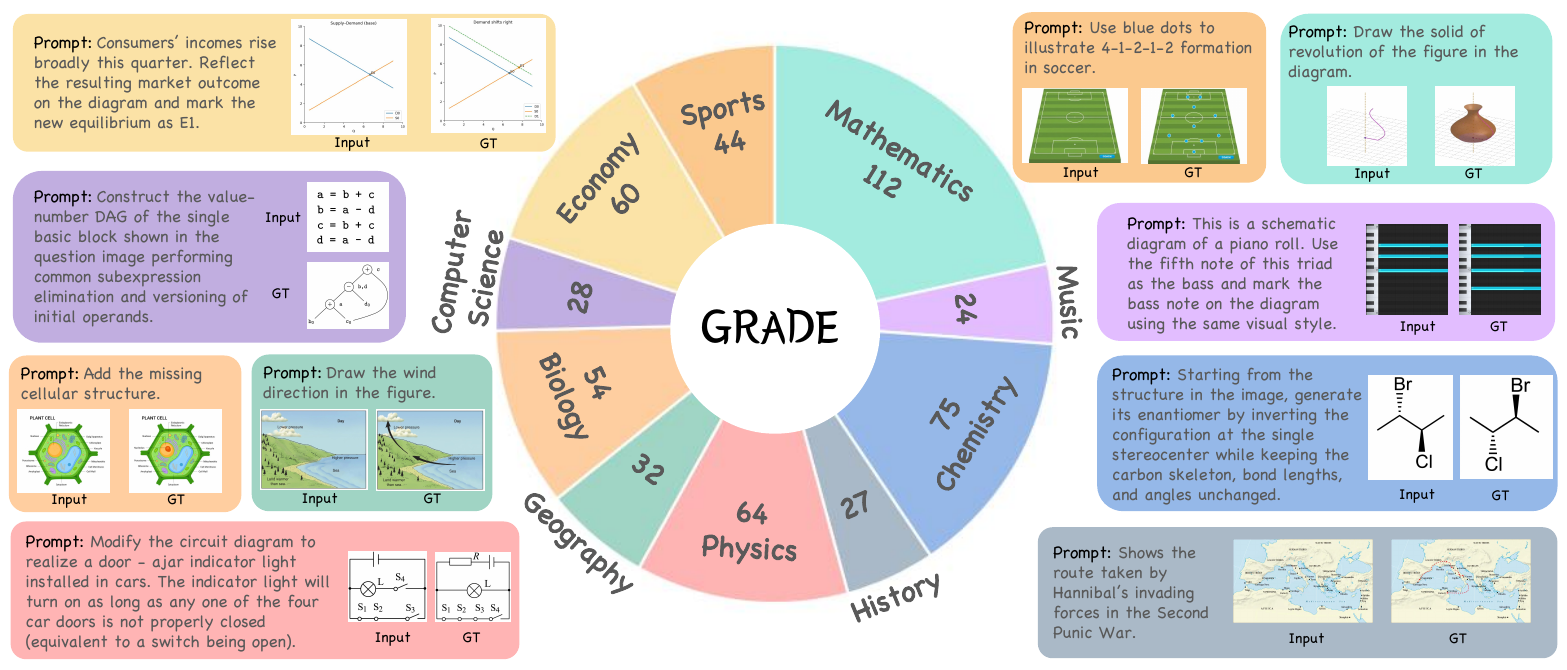}

    \caption{\textbf{Overview of \benchname.}  \benchname contains 520 discipline-informed image editing samples across ten academic disciplines. 
    }
    \label{fig:overview}
\end{figure*}
\section{\benchname Benchmark}

\subsection{Overview}



\noindent\textbf{Data Source.} 
Each sample in our dataset is organized as an image editing triplet, consisting of an input image, a textual editing instruction, and a corresponding GT image. Given the specialized and technically demanding nature of disciplinary knowledge, we adopt a carefully designed data construction and filtering pipeline to ensure both rigor and reliability.
For the majority of the dataset, six annotators with academic backgrounds in relevant disciplines source concept-grounded images from open textbooks, websites and other reference materials, then manually edit them to create input-GT pairs and design corresponding instructions, followed by cross-validation by two additional experts. For the remainder, we first apply an automated pipeline to coarsely filter suitable samples from MMMU~\cite{yue2024mmmu}, followed by manual curation where two experts select final samples and design editing instructions. The resulting data then undergoes cross-validation by two additional experts. Additional details on data collection are provided in the supplementary material.



\vspace{1mm}
\noindent\textbf{Taxonomy.} Fig.~\ref{fig:overview} illustrates the data distribution of our dataset along with representative examples. Our dataset covers 10 academic disciplines, including \textit{mathematics}, \textit{physics}, \textit{chemistry}, \textit{biology}, \textit{history}, \textit{geography}, \textit{sports}, \textit{music}, \textit{computer science}, and \textit{economics}. 
To further capture fine-grained knowledge structures, we introduce a hierarchical categorization scheme by defining second-level sub-disciplines within each primary domain.
For example, within the mathematics domain, we distinguish sub-disciplines such as \textit{plane geometry}, \textit{solid geometry}, \textit{functions}, \textit{graph} and \textit{statistics}, 
each targeting distinct forms of domain-specific knowledge and reasoning patterns. Similar hierarchical decompositions are applied across other disciplines to better reflect the diversity of academic concepts encountered in real-world editing scenarios. 
The detailed sub-discipline taxonomy for all ten disciplines is provided in the supplementary material.


\subsection{Evaluation Metrics}

\begin{figure*}[!tb]

    \centering
    \includegraphics[width=\linewidth]{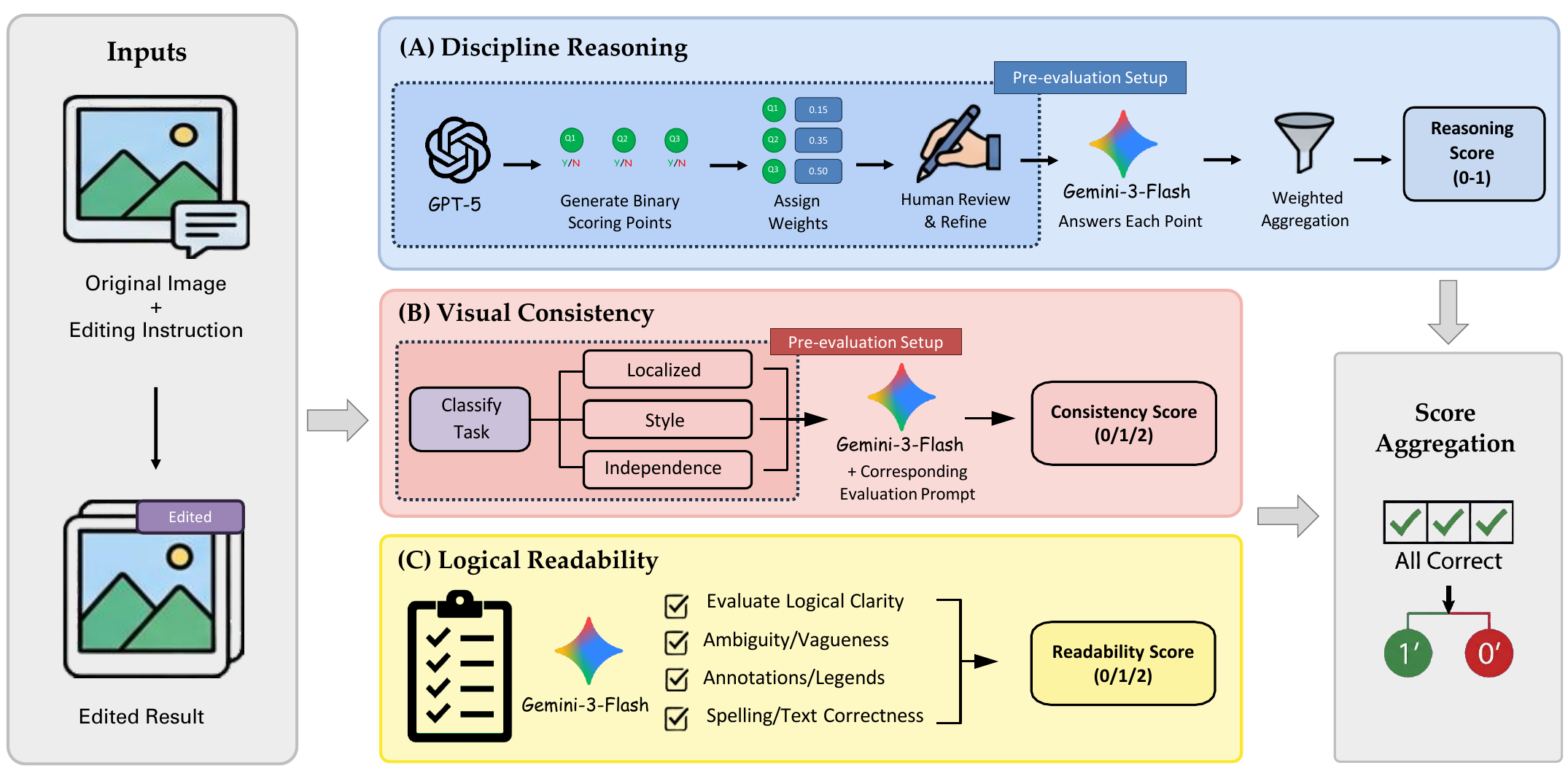}
    %
    \caption{\textbf{Evaluation pipeline.} We evaluate edited results on (A) \emph{Discipline Reasoning} via weighted, question-guided MLLM judging, (B) \emph{Visual Consistency} with task-specific prompts (localized/style/independence), and (C) \emph{Logical Readability} for clarity and text/annotation correctness. 
    }
    \label{fig:pipeline}
\end{figure*}

As the reasoning difficulty shifts toward complex discipline-informed knowledge, the design and rigor of evaluation become increasingly critical. To this end, we evaluate image editing performance along three dimensions: \textit{Discipline Reasoning}, \textit{Visual Consistency}, and \textit{Logical Readability}. 
Together, these complementary dimensions provide a comprehensive assessment of editing quality, covering both reasoning correctness and low-level visual quality. 
The overall evaluation pipeline is illustrated in Fig.~\ref{fig:pipeline}. 

\vspace{1mm}
\noindent\textbf{Discipline Reasoning.} We evaluate reasoning performance by assessing whether edited results correctly reflect the underlying disciplinary knowledge, which is central to the validity of discipline-informed image editing. Relying solely on human evaluation is costly and difficult to scale, while MLLM-as-a-judge~\cite{zhang2025large} approaches based on a single instruction often lead to unstable judgments. In practice, such prompts may inadequately capture all required criteria, resulting in high sensitivity to phrasing and limited reproducibility across evaluations.

Therefore, inspired by previous approaches~\cite{wang2025genexam, worldgenbench}, we adopt a structured, question-guided evaluation strategy. For each sample, we use GPT-5~\cite{singh2025openai} to generate weighted binary questions aligned with required disciplinary knowledge, with all weights summing to one. Two human experts with relevant disciplinary knowledge independently examine all scoring points, and the results are cross-validated by a third expert.
During evaluation, we use Gemini-3-Flash~\cite{Gemini-3-Flash} to assess the edited result by explicitly referencing these scoring points together with the GT image. 
The final reasoning score for each sample is computed as a weighted aggregation of the judge's responses, yielding a normalized score between 0 and 1.
By decomposing discipline-informed reasoning into explicit, verifiable criteria, the question-guided protocol enables targeted evaluation of whether models correctly understand and apply disciplinary concepts.

\vspace{1mm}
\noindent\textbf{Visual Consistency.} Visual consistency is a fundamental aspect of image editing quality, as it reflects whether an edit integrates coherently with the intended visual structure~\cite{zhao2025envisioning, wu2025kris}.
However, consistency requirements are inherently task-specific and vary across different editing scenarios.
To this end, we categorize consistency evaluation into three types: 

1. \textit{Localized Consistency}. This type applies to localized editing tasks, where only specific regions or elements are expected to change. 
For example, when completing missing entries in a timeline or adjusting curves in an economics diagram, all unrelated visual elements (\textit{i.e.} elements not mentioned in the editing instruction) should remain unchanged. Consistency is satisfied only if modifications are confined to the target components.

2. \textit{Style Consistency}. This type applies to global edits where the overall structure is modified, which can be viewed as style-consistent image-to-image generation. These tasks do not require local consistency of certain elements, but the visual representation style should be preserved.
For instance, when editing a chemical reaction diagram, the resulting molecule is expected to maintain the original bond-line representation rather than switching to alternative styles such as ball-and-stick models.

3. \textit{Consistency Independence}. This category applies to scenarios where consistency with the original image is not required, resembling general image-to-image generation. For example, given a rendered image of a mechanical part, the task may require generating its engineering orthographic views. In such cases, the edited output is expected to follow domain-specific representation standards rather than preserve visual consistency with the input image.

 The taxonomy was finalized through consensus among multiple annotators to minimize subjective bias. Based on these three categories, we design corresponding evaluation prompts to evaluate visual consistency. Each sample is assigned a consistency score of 0/1/2.

\vspace{1mm}
\noindent\textbf{Logical Readability}. Readability generally refers to whether an edited image can be clearly interpreted by a human viewer. 
Unlike natural image editing, where perceptual realism and visual sharpness are often sufficient, discipline-informed editing places stronger emphasis on whether knowledge is expressed in a logically coherent and structurally sound manner. An image that appears clear at a perceptual level may still fail to convey correct or interpretable academic meaning if its logical structure is flawed or its representations are ambiguous. For instance, curves in a diagram should be visually distinguishable, accompanied by clear annotations or legends, use correct and consistent textual labels, and follow a coherent representational convention to support reliable interpretation in academic contexts. 
Accordingly, we introduce a criterion for discipline-informed editing to assess whether the edited image presents discipline-specific content in a clear, logically consistent, and interpretable form. Each edited result is assigned a readability score of 0/1/2.

\vspace{1mm}
\noindent\textbf{Score Aggregation}. Inspired by prior benchmark designs, we adopt an overall accuracy that requires joint satisfaction of all evaluation criteria~\cite{zhao2025envisioning}. 
A sample is considered correct only if the model achieves the maximum score in all three dimensions; otherwise, it is counted as a failure. Additionally, we report a relaxed score that summarizes performance via criterion-wise aggregation~\cite{niu2025wiseworldknowledgeinformedsemantic}.
The relaxed score is computed as a weighted average of the three evaluation dimensions. We first normalize each dimension to range [0, 100], and assign weights of 0.6, 0.3, and 0.1 to Discipline Reasoning, Visual Consistency, and Logical Readability, respectively. 

\section{Experiment}
We conduct extensive experiments on our benchmark using a diverse set of 20 state-of-the-art models, including 10 closed-source models and 10 open-source models. The evaluated models cover both unified multimodal models, such as GPT-Image-1.5~\cite{GPT-Image-1.5} and Nano Banana~\cite{Gemini-3-Flash}, as well as specialized image editing models, including Qwen-Edit~\cite{wu2025qwenimagetechnicalreport} and FLUX~\cite{flux-2-2025}.
Our automated evaluation uses Gemini-3-Flash~\cite{Gemini-3-Flash} as the judge. 
We systematically analyze model performance across evaluation dimensions, with human alignment and ablation experiments to further interpret model behavior and evaluation reliability.

\begin{table*}[t]
\centering
\renewcommand{\arraystretch}{0.9}
\setlength{\tabcolsep}{4pt}
\caption{Performance comparison of different models across individual evaluation dimensions (normalized to 0-100) and the overall accuracy.
}
\label{main results}
\resizebox{\textwidth}{!}{
\begin{tabular}{lcccc}
\toprule
\textbf{Model} & \textbf{Reasoning} & \textbf{Consistency} & \textbf{Readability} & \textbf{Accuracy} \\
\midrule
\multicolumn{5}{l}{\textbf{$\blacktriangledown$ Closed Source Models}}\\
\midrule
Nano Banana Pro~\cite{Nano-Banana-Pro} & \textbf{77.5} & \textbf{89.5} & 95.8 & \textbf{46.2}\\
Nano Banana 2~\cite{Nano-Banana-2} & 72.6 & 86.4 & \textbf{95.9} & 39.6 \\
Seedream 5.0~\cite{Seedream5-0} & 64.1 & 87.5 & 90.6 & 24.7 \\
GPT-Image-1.5~\cite{GPT-Image-1.5} & 54.5 & 82.3 & 90.7 & 16.0 \\
FLUX.2 Max~\cite{flux-2-2025} & 47.8 & 67.2 & 68.6 & 11.9 \\
Nano Banana~\cite{team2023gemini} & 42.2 & 75.1 & 82.0 & 9.0 \\
Seedream 4.5~\cite{Seedream4-5} & 41.3 & 55.6 & 82.1 & 6.9 \\
GPT-Image-1.0~\cite{GPT-Image-1} & 44.0 & 65.2 & 82.3 & 6.0 \\
FLUX.2 Pro~\cite{flux-2-2025} & 38.9 & 55.5 & 70.3 & 4.4 \\
Seedream 4.0~\cite{seedream2025seedream} & 32.4 & 53.2 & 77.0 & 3.1 \\

\midrule
\multicolumn{5}{l}{\textbf{$\blacktriangledown$ Open Source Models}}\\
\midrule
Qwen-Edit-2511~\cite{wu2025qwenimagetechnicalreport} & 18.6 & 45.2 & 52.1 & 2.7 \\
Step-1X (think+reflect)~\cite{liu2025step1x} & 19.2 & 57.2 & 66.9 & 2.3 \\
FLUX.2 dev~\cite{flux-2-2025} & 23.0 & 56.4 & 69.0 & 2.1 \\
Step-1X (think)~\cite{liu2025step1x} & 17.6 & 56.3 & 68.2 & 1.4 \\
DreamOmni~\cite{xia2025dreamomniunifiedimagegeneration} & 17.4 & 83.2 & 89.1 & 1.0 \\
Step-1X~\cite{liu2025step1x} & 17.3 & 52.8 & 63.7 & 1.0 \\
Bagel~\cite{deng2025emerging} & 15.2 & 58.6 & 69.8 & 0.6 \\
Bagel (think)~\cite{deng2025emerging} & 15.6 & 54.8 & 67.8 & 0.2 \\
ICEdit~\cite{zhang2025icedit} & 9.8 & 33.2 & 56.5 & 0.2 \\
OmniGen~\cite{xiao2025omnigen} & 9.7 & 33.6 & 51.6 & 0.0 \\
\bottomrule
\end{tabular}
}
\end{table*}

\subsection{Main Result}


The performance of all evaluated models across individual dimensions and overall scores is summarized in Table~\ref{main results}. 
Among all the evaluated models, Nano Banana Pro~\cite{Nano-Banana-Pro} achieves the strongest overall performance, exhibiting clear advantages across all three dimensions with an accuracy of 46.2\%.
Notably, Nano Banana Pro and Nano Banana 2 significantly surpass other systems such as Seedream 5.0 (24.7\%). 
Despite this margin, current models can only attain an accuracy below 50\%, indicating that even the best-performing model fails to satisfy all discipline-informed editing requirements in more than half of the cases.
Interestingly, models such as Nano Banana Pro, GPT-Image-1.5, and Seedream 5.0, which achieve comparable performance on existing benchmarks~\cite{zhao2025envisioning,Artificial-Analysis}, exhibit substantial performance differences on GRADE (46.2\% vs. 16.0\% vs. 24.7\%), highlighting the stronger discriminative ability of GRADE in assessing knowledge-intensive reasoning in image editing.


Moreover, closed-source models consistently outperform open-source ones, where the best-performing open-source model, Qwen-Edit-2511 \cite{wu2025qwenimagetechnicalreport} (2.7\%), remains lower than all closed-source models, and most open-source models like OmniGen~\cite{xiao2025omnigen}, Bagel~\cite{deng2025emerging}, and FLUX.2 dev~\cite{flux-2-2025} only achieve near-zero or zero accuracy. As shown in Tab.~\ref{tab:model_performance_domain_rel_comp}, under the relaxed score, a similar performance gap between closed-source and open-source models is still observed, with most closed-source models scoring above 40\% while open-source models largely remain below this level.


\begin{table*}[t]
  \caption{Performance of different models across disciplines.}
  \centering
  \setlength{\tabcolsep}{3pt}
  \resizebox{\linewidth}{!}{
  \begin{tabular}{lcccccccccc}
  \toprule
  \textbf{Model} & \textbf{Phy} & \textbf{Sports} & \textbf{Chem} & \textbf{Math} & \textbf{Music} & \textbf{Econ} & \textbf{Hist} & \textbf{Geo} & \textbf{Bio} & \textbf{Comp} \\
  
  \midrule
  \multicolumn{11}{l}{\textbf{$\blacktriangledown$ Closed Source Models}} \\
  \midrule
  Nano Banana Pro~\cite{Nano-Banana-Pro} & \textbf{53.1} & \textbf{36.4} & 42.7 & \textbf{37.5} & \textbf{54.2} & 61.7 & \textbf{29.6} & \textbf{37.5} & \textbf{55.6} & \textbf{57.1} \\
  Nano Banana 2~\cite{Nano-Banana-2} & 35.9 & 31.8 & \textbf{44.0} & 33.9 & 37.5 & \textbf{71.7} & 22.2 & 21.9 & 38.9 & 42.9  \\
  Seedream 5.0~\cite{Seedream5-0} & 25.0 & 18.2 & 36.0 & 18.8 & 20.8 & 20.0 & 3.7 & 15.6 & 45.3 & 32.1  \\
  GPT-Image-1.5 & 15.6 & 13.6 & 24.0 & 9.8 & 4.2 & 20.0 & 0.0 & 25.0 & 22.2 & 17.9  \\
  FlUX.2 Max~\cite{flux-2-2025} & 3.1 & 9.1 & 21.3 & 7.1 & 12.5 & 3.3 & 3.7 & 12.5 & 29.6 & 21.4  \\
  Nano Banana~\cite{team2023gemini} & 1.6 & 4.6 & 12.0 & 6.3 & 16.7 & 15.0 & 3.7 & 9.4 & 13.0 & 14.3  \\
  Seedream 4.5~\cite{Seedream4-5} & 7.8 & 9.1 & 10.7 & 6.3 & 4.2 & 6.7 & 0.0 & 3.1 & 11.3 & 0.0  \\
  GPT-Image-1.0~\cite{GPT-Image-1} & 9.4 & 0.0 & 8.0 & 4.5 & 0.0 & 5.0 & 0.0 & 3.1 & 11.1 & 14.3  \\
  Flux.2 Pro~\cite{flux-2-2025} & 4.7 & 4.6 & 6.7 & 4.5 & 0.0 & 1.7 & 0.0 & 6.3 & 5.6 & 7.1  \\
  Seedream 4.0~\cite{seedream2025seedream} & 0.0 & 2.3 & 5.3 & 4.5 & 4.2 & 3.3 & 0.0 & 3.1 & 3.8 & 0.0  \\
  \midrule
  \multicolumn{11}{l}{\textbf{$\blacktriangledown$ Open Source Models}} \\
  \midrule
  Qwen-Edit-2511~\cite{wu2025qwenimagetechnicalreport} & 0.0 & 0.0 & 0.0 & 9.8 & 0.0 & 0.0 & 0.0 & 6.3 & 0.0 & 3.6 \\
  Step-1X~\cite{liu2025step1x} (think+reflect) & 0.0 & 0.0 & 1.3 & 1.8 & 8.3 & 1.7 & 7.1 & 0.0 & 7.4 & 0.0 \\
  FLUX.2 dev~\cite{flux-2-2025} & 0.0 & 2.3 & 1.3 & 3.6 & 4.2 & 0.0 & 0.0 & 3.1 & 5.6 & 0.0 \\
  Step-1X~\cite{liu2025step1x} (think) & 0.0 & 0.0 & 2.7 & 0.0 & 4.2 & 0.0 & 3.7 & 6.3 & 1.9 & 0.0 \\
  DreamOmni~\cite{xia2025dreamomniunifiedimagegeneration} & 0.0 & 0.0 & 1.3 & 0.0 & 4.2 & 1.7 & 0.0 & 0.0 & 3.7 & 0.0 \\
  Step-1X~\cite{liu2025step1x} & 1.6 & 0.0 & 0.0 & 0.0 & 0.0 & 0.0 & 3.7 & 6.3 & 0.0 & 0.0 \\
  Bagel & 0.0 & 0.0 & 1.3 & 0.0 & 4.2 & 0.0 & 0.0 & 0.0 & 1.9 & 0.0 \\
  Bagel~\cite{deng2025emerging} (think)  & 0.0 & 0.0 & 0.0 & 0.9 & 0.0 & 0.0 & 0.0 & 0.0 & 0.0 & 0.0 \\
  ICEdit~\cite{zhang2025icedit} & 0.0 & 0.0 & 0.0 & 0.0 & 0.0 & 1.7 & 0.0 & 0.0 & 0.0 & 0.0 \\
  OmniGen~\cite{xiao2025omnigen}  & 0.0 & 0.0 & 0.0 & 0.0 & 0.0 & 0.0 & 0.0 & 0.0 & 0.0 & 0.0 \\
  \bottomrule
  \end{tabular}
  }
  \label{tab:model_performance_domain}
\end{table*}
\begin{table*}[h]
\caption{Relaxed Score of different models across subjects. 
}
\centering
\setlength{\tabcolsep}{4pt}
\resizebox{\linewidth}{!}{
\begin{tabular}{lccccccccccc}
\toprule
\textbf{Model} & \textbf{Phy} & \textbf{Sports} & \textbf{Chem} & \textbf{Math} & \textbf{Music} & \textbf{Econ} & \textbf{Hist} & \textbf{Geo} & \textbf{Bio} & \textbf{Comp} & \makecell{\textbf{Relaxed}\\ \textbf{Score}} \\
\midrule

\multicolumn{12}{l}{\textbf{$\blacktriangledown$ Closed Source Models}} \\
\midrule

Nano Banana Pro~\cite{Nano-Banana-Pro} & \textbf{84.0} & \textbf{75.7} & \textbf{85.6} & \textbf{75.9} & \textbf{82.8} & 89.5 & \textbf{81.7} & \textbf{80.5} & \textbf{89.0} & \textbf{90.1} & \textbf{82.9} \\
Nano Banana 2~\cite{Nano-Banana-2} & 77.2 & 68.8 & 83.2 & 74.7 & 76.7 & \textbf{92.6} & 75.5 & 76.2 & 83.2 & 78.1 & 79.1 \\
Seedream 5.0~\cite{Seedream5-0} & 74.1 & 72.9 & 79.7 & 67.4 & 62.6 & 79.5 & 66.2 & 71.5 & 82.6 & 74.8 & 73.7 \\
GPT-Image-1.5~\cite{GPT-Image-1.5} & 60.8 & 67.8 & 72.9 & 58.5 & 62.3 & 72.1 & 53.8 & 77.1 & 74.2 & 69.0 & 66.5 \\
Nano Banana~\cite{team2023gemini} & 56.9 & 58.1 & 58.7 & 49.3 & 44.5 & 59.5 & 54.3 & 56.7 & 63.4 & 60.5 & 56.0 \\
FLUX.2 Max~\cite{flux-2-2025} & 54.7 & 56.0 & 63.2 & 45.4 & 38.6 & 53.0 & 64.2 & 60.6 & 69.7 & 57.8 & 55.7 \\
GPT-Image-1.0~\cite{GPT-Image-1} & 55.4 & 56.7 & 61.9 & 43.3 & 51.5 & 57.8 & 44.5 & 58.4 & 60.2 & 58.0 & 54.2 \\
Seedream 4.5~\cite{Seedream4-5} & 53.6 & 54.4 & 57.4 & 40.5 & 40.7 & 55.8 & 35.6 & 43.2 & 60.2 & 44.9 & 49.7 \\
FLUX.2 Pro & 46.6 & 54.0 & 50.6 & 43.3 & 28.0 & 36.0 & 43.7 & 55.2 & 59.3 & 53.7 & 47.0 \\
Seedream 4.0~\cite{seedream2025seedream} & 41.8 & 51.8 & 48.1 & 36.0 & 32.4 & 49.3 & 28.9 & 41.7 & 49.3 & 47.2 & 43.1 \\

\midrule
\multicolumn{12}{l}{\textbf{$\blacktriangledown$ Open Source Models}} \\
\midrule

DreamOmni~\cite{xia2025dreamomniunifiedimagegeneration} & 40.6 & 52.7 & 50.6 & 40.1 & 47.4 & 44.9 & 35.5 & 35.9 & 47.0 & 48.9 & 44.3 \\
FLUX.2 dev~\cite{flux-2-2025} & 34.3 & 54.3 & 34.0 & 37.0 & 34.8 & 32.8 & 40.4 & 42.9 & 37.1 & 36.7 & 37.6 \\
Step-1X~\cite{liu2025step1x} (t+r) & 31.1 & 45.3 & 33.6 & 30.6 & 32.1 & 29.2 & 54.3 & 44.8 & 43.1 & 25.6 & 35.4 \\
Step-1X~\cite{liu2025step1x} (think) & 30.9 & 45.0 & 34.0 & 29.3 & 31.7 & 27.5 & 49.8 & 48.7 & 38.1 & 23.7 & 34.3 \\
Bagel~\cite{deng2025emerging} & 27.7 & 44.3 & 35.4 & 34.3 & 34.4 & 27.2 & 38.5 & 35.2 & 34.0 & 30.7 & 33.7 \\
Step-1X~\cite{liu2025step1x} & 30.2 & 41.8 & 27.0 & 27.9 & 32.0 & 29.3 & 50.8 & 45.8 & 35.0 & 27.5 & 32.6 \\
Bagel~\cite{deng2025emerging} (think) & 28.2 & 43.5 & 32.2 & 29.9 & 35.4 & 28.7 & 41.7 & 25.1 & 34.8 & 37.4 & 32.6 \\
Qwen-edit-2511~\cite{wu2025qwenimagetechnicalreport} & 27.7 & 45.1 & 13.7 & 39.5 & 19.8 & 28.6 & 36.9 & 33.0 & 27.5 & 23.9 & 30.0 \\
ICEdit~\cite{zhang2025icedit} & 18.9 & 37.1 & 18.6 & 19.5 & 30.6 & 22.9 & 12.3 & 24.9 & 19.6 & 16.3 & 21.5 \\
OmniGen~\cite{xiao2025omnigen} & 19.5 & 44.0 & 15.7 & 15.3 & 30.0 & 11.5 & 24.8 & 23.4 & 21.2 & 31.8 & 21.1 \\

\bottomrule
\end{tabular}
}
\label{tab:model_performance_domain_rel_comp}
\end{table*}
From a dimension-wise perspective, the results reveal clear and consistent performance disparities across models. Closed-source models substantially outperform open-source ones: Nano Banana Pro achieves 77.5\% in Reasoning, markedly higher than Seedream 5.0 (64.1\%) and GPT-Image-1.5 (54.5\%), while the best open-source model, Qwen-Edit-2511, reaches only 18.6\%. Notably, even among closed-source models, reasoning ability varies significantly, with Nano Banana Pro exceeding Seedream 5.0 by 13.4\%, indicating that discipline-informed editing exposes fine-grained differences in academic reasoning capability. In contrast, Visual Consistency and Logical Readability function as constraint-oriented criteria rather than ranking-oriented metrics. While many models achieve moderate scores on these dimensions, they play a critical diagnostic role by exposing distinct failure modes. For example, FLUX.2 dev exhibits a notably low consistency score (17.6\%), indicating severe violations of instruction-scoped editing constraints and unintended structural modifications. Conversely, although DreamOmni~\cite{xia2025dreamomniunifiedimagegeneration} attains relatively high Visual Consistency scores (83.2\%) and Logical Readability scores (89.1\%), its overall performance remains low, largely because the model favors minimal or no changes to preserve the original image. These two dimensions complement Discipline Reasoning by enforcing structural and representational constraints, ensuring that high reasoning scores reflect valid and interpretable edits rather than superficial solutions.

As shown in Tab.~\ref{tab:model_performance_domain}, across disciplines, clear performance disparities emerge. Nano Banana Pro consistently achieves the highest scores across nearly all disciplines,
indicating robust and stable discipline-informed reasoning rather than isolated domain strengths. Specifically, STEM-oriented disciplines such as Physics, Biology, and Mathematics exhibit the strongest differentiation: for example, in Physics, Nano Banana Pro achieves 53.1\%, notably higher than Seedream 5.0 (25.0\%) and GPT-Image-1.5 (15.6\%), with similar gaps observed in Biology (55.6\% vs. 45.3\% and 22.2\%, respectively). In contrast, humanities-related disciplines, particularly History and Geography, remain challenging even for closed-source models, with accuracy dropping sharply (e.g., Nano Banana Pro attains only 29.6\%). Meanwhile, open-source models largely fail across disciplines, with accuracy frequently at or near zero. 





\subsection{Human Alignment}
\label{sec:exp_human}
We analyze the alignment between our automated evaluation pipeline and human judgment on 68 samples uniformly sampled across disciplines, using outputs from three representative models (Nano Banana Pro, GPT-Image-1.5, Qwen-Edit-2511), where results are averaged across models and examined separately for each evaluation dimension.
Human experts are required to follow the evaluation protocol to rate the generated images of Discipline Reasoning, Visual Consistency and Logical Readability.
We collect annotations from five human experts and use the averaged human scores to mitigate individual bias. The alignment between human judgments and automated scores is measured using both mean absolute error (MAE) and standard deviation (STD), all values are normalized to the [0, 1] range. As shown in the last row of Tab.~\ref{tab:MAE}, the MAE values across all three dimensions are around 10\% while the STD remains relatively stable across dimensions, indicating consistent variance in scoring deviations.

\begin{table}[htbp]
    \centering
    \caption{\textbf{Alignment between automated evaluation and human judgments.} We report mean absolute error (MAE) and standard deviation (STD) of three evaluation dimensions.
    }
    \resizebox{0.9\textwidth}{!}{
    \begin{tabular}{lcccccc}
      \toprule
      \multirow{2}{*}{\textbf{Judge Model}} & \multicolumn{2}{c}{\textbf{Reasoning}} & \multicolumn{2}{c}{\textbf{Consistency}} & \multicolumn{2}{c}{\textbf{Readability}} \\
      \cmidrule(lr){2-3} \cmidrule(lr){4-5} \cmidrule(lr){6-7} 
        & \textbf{MAE} ($\downarrow$) & \textbf{STD} ($\downarrow$) & \textbf{MAE} ($\downarrow$) & \textbf{STD} ($\downarrow$) & \textbf{MAE} ($\downarrow$) & \textbf{STD} ($\downarrow$)  \\
      \midrule

      Qwen3-VL-235B  & 0.1798 & 0.3504 & 0.1231 & 0.3523 & 0.0838 & 0.2944  \\
      GPT-5 & 0.1519 & 0.3227 & 0.1677 & 0.4386 & 0.1221 & 0.3638   \\
      \textbf{Gemini-3-Flash} & \textbf{0.1194} & \textbf{0.2834} & \textbf{0.0954} & \textbf{0.3047} & \textbf{0.0838} & \textbf{0.2944}   \\
      
      \bottomrule
    \end{tabular}
    }
    
    \label{tab:MAE}
\end{table}

\subsection{Ablation Study}

\textbf{Judge Model}. To study the impact of the judging model, we compare several commonly used MLLMs for automated evaluation, including an open-source model (Qwen3-VL-235B-Instruct~\cite{Qwen3-VL}) and a closed-source model (GPT-5~\cite{singh2025openai}), against our default judge, Gemini-3-Flash~\cite{Gemini-3-Flash}. All judges are evaluated on the same 68-sample subset by measuring their alignment with human annotations in Sec.~\ref{sec:exp_human}.
As shown in Tab.~\ref{tab:MAE}, Gemini-3-Flash consistently achieves the lowest MAE across all three evaluation dimensions, indicating the strongest alignment with human judgments. It also exhibits the lowest STD, reflecting more stable and consistent scoring behavior across samples. These results indicate that Gemini-3-Flash provides a reliable approximation of human evaluation and is well-suited for use as the judging model in our automated evaluation pipeline.


\vspace{1mm}
\noindent\textbf{Instruction Explicitness}. All instructions in our benchmark are designed to require implicit reasoning. For example, an instruction may ask for the product of a chemical reaction given a molecular structure, without explicitly describing how the structure should be transformed. To study the role of instruction formulation, we convert all instructions into explicit versions that directly specify the required knowledge-driven edits, while keeping the input and ground truth images unchanged. 
We select one open-source model (Qwen-Edit-2511) and one closed-source model (Nano Banana 2) with moderate overall performance on the full benchmark for this ablation study. 
Their performance under implicit and explicit instructions on the 68-sample subset is reported in Tab.~\ref{tab:explicit}.
As shown, making the instructions explicit consistently improves performance for both models. The largest gains among the three metrics are observed in Discipline Reasoning, where Nano Banana 2 improves from 67.9\% to 89.7\% and Qwen-Edit-2511 from 18.9\% to 44.7\%. Overall, the open-source model Qwen-Edit-2511 exhibits a larger relative improvement, with accuracy increasing from 1.5\% to 8.8\%, suggesting that open-source models rely more on explicit guidance than closed-source models, reflecting a larger gap in implicit reasoning capability.

\begin{table*}
    \centering
    
    \caption{Ablation study on instruction explicitness.}
    \resizebox{0.8\textwidth}{!}{ 
    \begin{tabular}{l |c c c c}
    \toprule
     \textbf{Model} & \textbf{Reasoning} & \textbf{Consistency} & \textbf{Readability} & \textbf{Accuracy}   \\
     \hline
     Nano Banana 2 & 67.9 & 83.8 & 95.6 & 35.3\\
     \ \ w/ explicit inst. & 89.7 & 93.0 & 95.5 & 65.7  \\
     \hline
     Qwen-Edit-2511 & 18.9 & 40.0 & 47.8 & 1.5  \\
     \ \ w/ explicit inst. & 44.7 & 60.0 & 81.6 & 8.8 \\
    \bottomrule
    \end{tabular}
    }
    \label{tab:explicit}
\end{table*}

\subsection{Case Study}

\begin{figure*}[tb!]

    \centering
    
    \includegraphics[width=\linewidth]{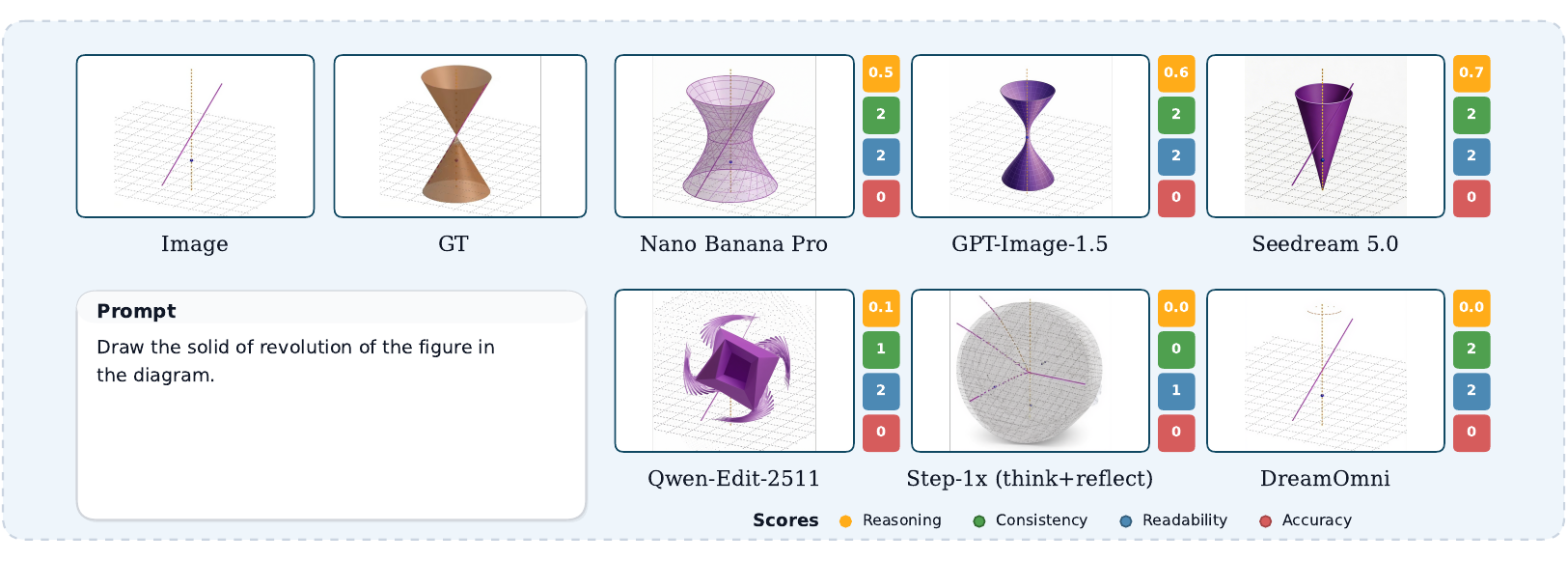}
    \includegraphics[width=\linewidth]{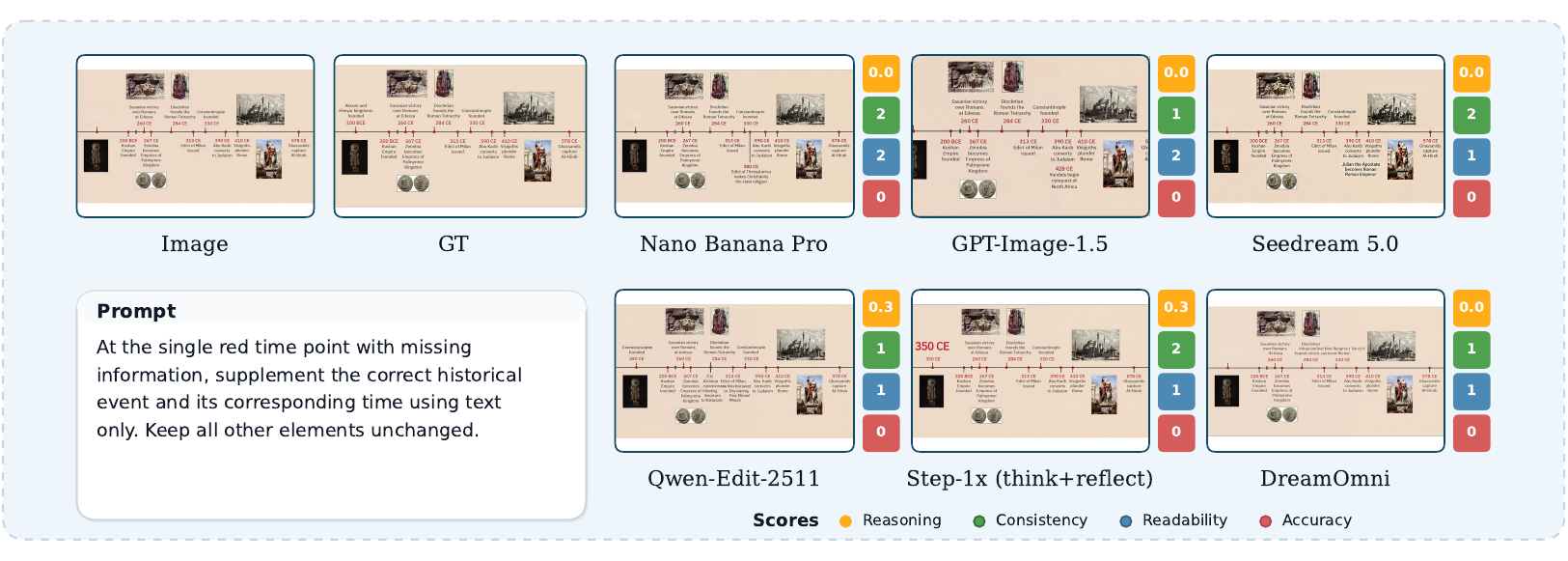}
    
   
    \caption{Qualitative comparison of six representative high-performing models. 
    }
    \label{fig:case}
\end{figure*}

As shown in Fig.~\ref{fig:case}, we present a qualitative case study of six representative models that achieve relatively strong overall performance on the benchmark, including three closed-source (Nano Banana Pro, GPT-Image-1.5 and Seedream 5.0) and three open-source models (Qwen-Edit-2511, Step-1X (think+reflect) and DreamOmni).



For the example above, the diagram suggests that both Nano Banana Pro and GPT-Image-1.5 demonstrate an understanding of the concepts of the generatrix and the axis of rotation. Based on these cues, they further infer that the resulting shape should exhibit the characteristic geometry of being narrow at the top and bottom while wider in the middle. However, neither model is able to precisely reason about the spatial relationship between the generatrix and the rotation axis, and thus fails to conclude that the correct result should be an hourglass-shaped solid composed of two standard cones. Meanwhile, Seedream 5.0 correctly recognizes the concept of a solid of revolution but incorrectly infers the resulting shape to be a single cone. The remaining three open-source models fail to demonstrate an understanding of solids generated by rotation in three-dimensional space, indicating a clear limitation in spatial reasoning related to rotational geometry.

The other example further challenges the model's integrated reasoning capabilities. The model must first identify the missing text region, infer the underlying historical event, and then insert the corresponding textual content. Interestingly, we observe an unexpected phenomenon: none of the strong closed-source models correctly identified the missing position. 
Instead, they tended to insert additional information into other available empty spaces in the figure. 
In contrast, some of the open-source models (e.g., Qwen-Edit-2511 and Step-1X (think+reflect)) were able to correctly infer where the missing text should be placed, yet failed at the knowledge reasoning stage, leading to the incorrect identification of the corresponding year and historical event.


\begin{figure}[tb!]
    \centering 
   
    \includegraphics[width=1\linewidth]{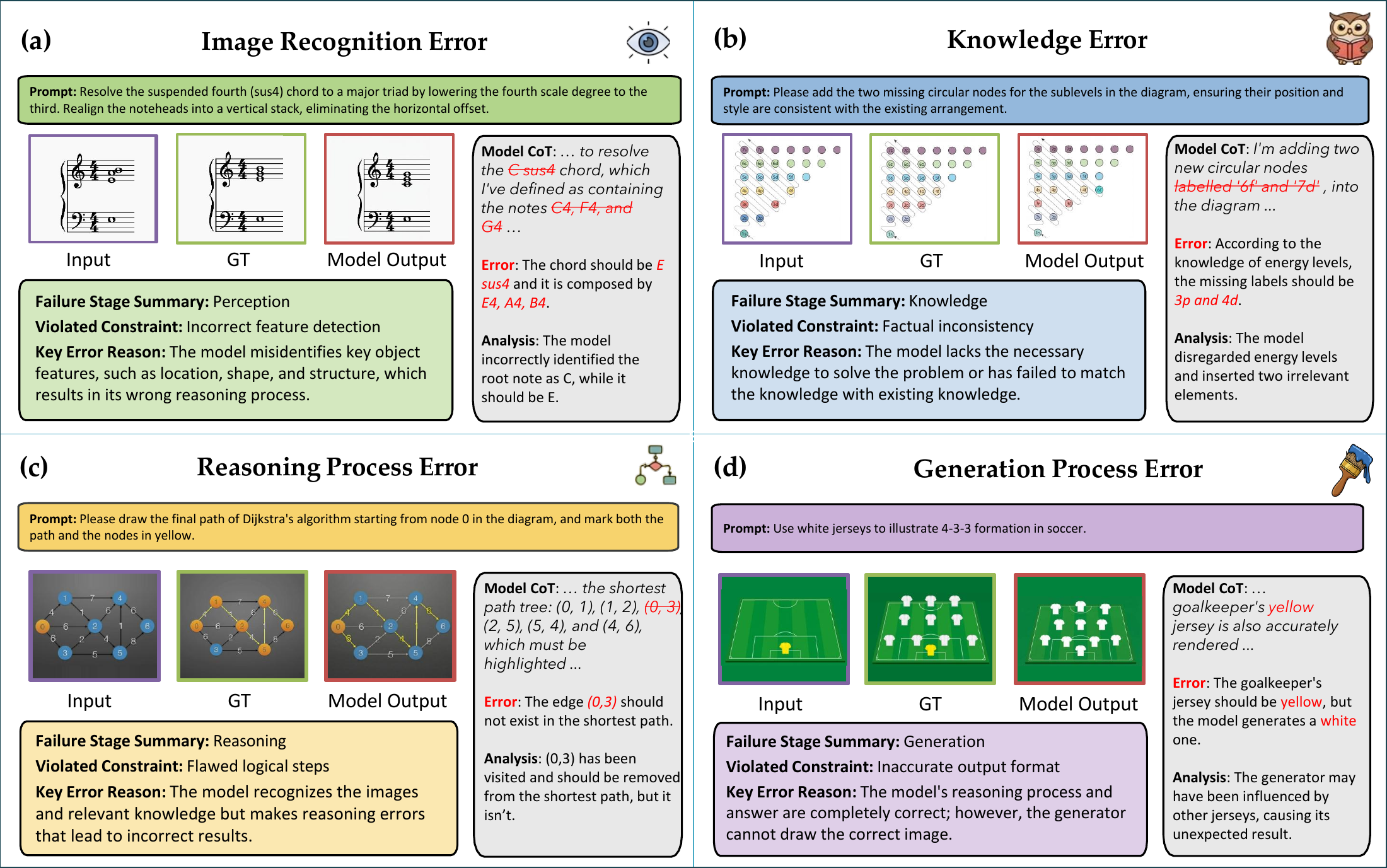}

    \caption{\textbf{Error analysis of Nano Banana Pro.} GT: Ground truth image. Error types include: (a) Image recognition error: mis-parsed structural cues. (b) Knowledge error: failure to activate domain-specific priors produces semantically invalid elements. (c) Reasoning process error: correct methodology but flawed multi-step execution violates constraints. (d) Generation process error: correct planning but failure to enforce hard constraints during synthesis.}
    \label{fig:Error_Analysis}

\end{figure}

\subsection{Error Analysis}


Testing models' knowledge-based reasoning is the core of our benchmark.
As illustrated in Fig.~\ref{fig:Error_Analysis}, we conduct a targeted error analysis of Nano Banana Pro to examine why current SoTA models still fall short, jointly analyzing its final outputs and chain-of-thought process.
This enables us to pinpoint systematic bottlenecks across perception, knowledge grounding, procedural execution, and synthesis, revealing whether failures stem from input misperception, inadequate knowledge retrieval, breakdowns in multi-step reasoning, or constraint violations during generation.
We categorize them into four representative error types: 

\noindent\textbf{(1) Image Recognition Error.} The model fails to robustly extract what each symbol is, where it is, and how elements align/connect in dense, structured visuals, so downstream reasoning is grounded on a wrong perceptual parse. 

\noindent\textbf{(2) Knowledge Error.} When an image follows a discipline-specific canonical template (e.g., orbital energy diagrams, reaction-mechanism arrows, chart axes semantics, anatomical labels, geometric construction rules), failing to trigger the relevant priors makes the model treat the task as generic "shape completion/node adding," producing elements that are semantically invalid for the system. 

\noindent\textbf{(3) Reasoning Process Error.} The model correctly recognizes entities and identifies the right methodology (e.g., shortest paths, conservation laws, grammar rules, geometric theorems), yet deviates during multi-step execution by missing constraints, state updates, or selection criteria, treating intermediate artifacts as final results. 

\noindent\textbf{(4) Generation Process Error.} The model can correctly plan the target and constraints ("what to keep" and "which attributes must remain fixed"), but in final image synthesis, these are not reliably enforced at low-level control, so the output drifts due to priors, global style consistency, or pattern copying. 

\section{Conclusion}

In this work, we present \benchname, a discipline-informed image editing benchmark that systematically evaluates editing performance grounded in discipline-informed knowledge across ten academic disciplines. 
\benchname comprises 520 carefully constructed samples of 10 disciplines and provides a multi-dimensional evaluation protocol that demonstrates strong alignment with human judgments. 
Through extensive experiments on a wide range of models, our benchmark reveals substantial performance gaps, particularly in applying discipline-informed reasoning under implicit instructions. These results highlight persistent limitations of current models in handling structured academic knowledge beyond visual realism, and point to the need for future unified multimodal models to better integrate discipline-informed knowledge, reasoning, and editing.



\section*{Acknowledgement}
This work was supported by the National Natural Science Foundation of China (62506229), Natural Science Foundation of Shanghai under 25ZR1402268, and Shanghai QiYuan Innovation Foundation.
%
%
\clearpage
\bibliographystyle{splncs04}
\bibliography{main}

@article{ho2020denoising,
  title={Denoising diffusion probabilistic models},
  author={Ho, Jonathan and Jain, Ajay and Abbeel, Pieter},
  journal={Advances in neural information processing systems},
  volume={33},
  pages={6840--6851},
  year={2020}
}

@article{dhariwal2021diffusion,
  title={Diffusion models beat gans on image synthesis},
  author={Dhariwal, Prafulla and Nichol, Alexander},
  journal={Advances in neural information processing systems},
  volume={34},
  pages={8780--8794},
  year={2021}
}

@inproceedings{crowson2022vqgan,
  title={Vqgan-clip: Open domain image generation and editing with natural language guidance},
  author={Crowson, Katherine and Biderman, Stella and Kornis, Daniel and Stander, Dashiell and Hallahan, Eric and Castricato, Louis and Raff, Edward},
  booktitle={European conference on computer vision},
  pages={88--105},
  year={2022},
  organization={Springer}
}

@article{balaji2022ediff,
  title={ediff-i: Text-to-image diffusion models with an ensemble of expert denoisers},
  author={Balaji, Yogesh and Nah, Seungjun and Huang, Xun and Vahdat, Arash and Song, Jiaming and Zhang, Qinsheng and Kreis, Karsten and Aittala, Miika and Aila, Timo and Laine, Samuli and others},
  journal={arXiv preprint arXiv:2211.01324},
  year={2022}
}

@inproceedings{zhang2023adding,
  title={Adding conditional control to text-to-image diffusion models},
  author={Zhang, Lvmin and Rao, Anyi and Agrawala, Maneesh},
  booktitle={Proceedings of the IEEE/CVF international conference on computer vision},
  pages={3836--3847},
  year={2023}
}

@article{saharia2022photorealistic,
  title={Photorealistic text-to-image diffusion models with deep language understanding},
  author={Saharia, Chitwan and Chan, William and Saxena, Saurabh and Li, Lala and Whang, Jay and Denton, Emily L and Ghasemipour, Kamyar and Gontijo Lopes, Raphael and Karagol Ayan, Burcu and Salimans, Tim and others},
  journal={Advances in neural information processing systems},
  volume={35},
  pages={36479--36494},
  year={2022}
}

@inproceedings{li2025synergen,
  title={Synergen-vl: Towards synergistic image understanding and generation with vision experts and token folding},
  author={Li, Hao and Tian, Changyao and Shao, Jie and Zhu, Xizhou and Wang, Zhaokai and Zhu, Jinguo and Dou, Wenhan and Wang, Xiaogang and Li, Hongsheng and Lu, Lewei and others},
  booktitle={Proceedings of the Computer Vision and Pattern Recognition Conference},
  pages={29767--29779},
  year={2025}
}

@article{zhang2025unimodel,
  title={UniModel: A Visual-Only Framework for Unified Multimodal Understanding and Generation},
  author={Zhang, Chi and Wang, Jiepeng and Wang, Youming and Liang, Yuanzhi and Yang, Xiaoyan and Li, Zuoxin and Huang, Haibin and Li, Xuelong},
  journal={arXiv preprint arXiv:2511.16917},
  year={2025}
}

@article{team2024chameleon,
  title={Chameleon: Mixed-modal early-fusion foundation models, 2024},
  author={Team, Chameleon},
  journal={URL https://arxiv. org/abs/2405.09818},
  volume={9},
  number={8},
  year={2024}
}

@article{deng2025emerging,
  title={Emerging properties in unified multimodal pretraining},
  author={Deng, Chaorui and Zhu, Deyao and Li, Kunchang and Gou, Chenhui and Li, Feng and Wang, Zeyu and Zhong, Shu and Yu, Weihao and Nie, Xiaonan and Song, Ziang and others},
  journal={arXiv preprint arXiv:2505.14683},
  year={2025}
}

@article{liu2025step1x,
  title={Step1x-edit: A practical framework for general image editing},
  author={Liu, Shiyu and Han, Yucheng and Xing, Peng and Yin, Fukun and Wang, Rui and Cheng, Wei and Liao, Jiaqi and Wang, Yingming and Fu, Honghao and Han, Chunrui and others},
  journal={arXiv preprint arXiv:2504.17761},
  year={2025}
}

@inproceedings{xiao2025omnigen,
  title={Omnigen: Unified image generation},
  author={Xiao, Shitao and Wang, Yueze and Zhou, Junjie and Yuan, Huaying and Xing, Xingrun and Yan, Ruiran and Li, Chaofan and Wang, Shuting and Huang, Tiejun and Liu, Zheng},
  booktitle={Proceedings of the Computer Vision and Pattern Recognition Conference},
  pages={13294--13304},
  year={2025}
}

@misc{wu2025qwenimagetechnicalreport,
      title={Qwen-Image Technical Report}, 
      author={Chenfei Wu and Jiahao Li and Jingren Zhou and Junyang Lin and Kaiyuan Gao and Kun Yan and Sheng-ming Yin and Shuai Bai and Xiao Xu and Yilei Chen and Yuxiang Chen and Zecheng Tang and Zekai Zhang and Zhengyi Wang and An Yang and Bowen Yu and Chen Cheng and Dayiheng Liu and Deqing Li and Hang Zhang and Hao Meng and Hu Wei and Jingyuan Ni and Kai Chen and Kuan Cao and Liang Peng and Lin Qu and Minggang Wu and Peng Wang and Shuting Yu and Tingkun Wen and Wensen Feng and Xiaoxiao Xu and Yi Wang and Yichang Zhang and Yongqiang Zhu and Yujia Wu and Yuxuan Cai and Zenan Liu},
      year={2025},
      eprint={2508.02324},
      archivePrefix={arXiv},
      primaryClass={cs.CV},
      url={https://arxiv.org/abs/2508.02324}, 
}

@inproceedings{zhang2025icedit,
  title     = {In-Context Edit: Enabling Instructional Image Editing with In-Context Generation in Large-Scale Diffusion Transformers},
  author    = {Zhang, Zechuan and Xie, Ji and Lu, Yu and Yang, Zongxin and Yang, Yi},
  booktitle = {Advances in Neural Information Processing Systems (NeurIPS)},
  year      = {2025},
  note      = {arXiv:2504.20690}
}

@misc{flux-2-2025,
    author={Black Forest Labs},
    title={{FLUX.2: Frontier Visual Intelligence}},
    year={2025},
    howpublished={\url{https://bfl.ai/blog/flux-2}},
}

@article{singh2025openai,
  title={Openai gpt-5 system card},
  author={Singh, Aaditya and Fry, Adam and Perelman, Adam and Tart, Adam and Ganesh, Adi and El-Kishky, Ahmed and McLaughlin, Aidan and Low, Aiden and Ostrow, AJ and Ananthram, Akhila and others},
  journal={arXiv preprint arXiv:2601.03267},
  year={2025}
}

@article{team2023gemini,
  title={Gemini: a family of highly capable multimodal models},
  author={Team, Gemini and Anil, Rohan and Borgeaud, Sebastian and Alayrac, Jean-Baptiste and Yu, Jiahui and Soricut, Radu and Schalkwyk, Johan and Dai, Andrew M and Hauth, Anja and Millican, Katie and others},
  journal={arXiv preprint arXiv:2312.11805},
  year={2023}
}

@article{seedream2025seedream,
  title={Seedream 4.0: Toward next-generation multimodal image generation},
  author={Seedream, Team and Chen, Yunpeng and Gao, Yu and Gong, Lixue and Guo, Meng and Guo, Qiushan and Guo, Zhiyao and Hou, Xiaoxia and Huang, Weilin and Huang, Yixuan and others},
  journal={arXiv preprint arXiv:2509.20427},
  year={2025}
}

@misc{Seedream4-5,
  title={Seedream 4.5},
  author={Seedream},
  howpublished={\url{https://seed.bytedance.com/en/seedream4_5/}},
  year={2025}
}

@misc{Seedream5-0,
  title={Seedream 5.0},
  author={Seedream},
  howpublished={\url{https://seed.bytedance.com/en/seedream5_0_lite/}},
  year={2026}
}

@misc{xia2025dreamomniunifiedimagegeneration,
      title={DreamOmni: Unified Image Generation and Editing}, 
      author={Bin Xia and Yuechen Zhang and Jingyao Li and Chengyao Wang and Yitong Wang and Xinglong Wu and Bei Yu and Jiaya Jia},
      year={2025},
      eprint={2412.17098},
      archivePrefix={arXiv},
      primaryClass={cs.CV},
      url={https://arxiv.org/abs/2412.17098}, 
}

@inproceedings{yue2024mmmu,
  title={Mmmu: A massive multi-discipline multimodal understanding and reasoning benchmark for expert agi},
  author={Yue, Xiang and Ni, Yuansheng and Zhang, Kai and Zheng, Tianyu and Liu, Ruoqi and Zhang, Ge and Stevens, Samuel and Jiang, Dongfu and Ren, Weiming and Sun, Yuxuan and others},
  booktitle={Proceedings of the IEEE/CVF Conference on Computer Vision and Pattern Recognition},
  pages={9556--9567},
  year={2024}
}

@article{phan2025humanity,
  title={Humanity's last exam},
  author={Phan, Long and Gatti, Alice and Han, Ziwen and Li, Nathaniel and Hu, Josephina and Zhang, Hugh and Zhang, Chen Bo Calvin and Shaaban, Mohamed and Ling, John and Shi, Sean and others},
  journal={arXiv preprint arXiv:2501.14249},
  year={2025}
}

@article{zhao2025envisioning,
  title={Envisioning beyond the pixels: Benchmarking reasoning-informed visual editing},
  author={Zhao, Xiangyu and Zhang, Peiyuan and Tang, Kexian and Zhu, Xiaorong and Li, Hao and Chai, Wenhao and Zhang, Zicheng and Xia, Renqiu and Zhai, Guangtao and Yan, Junchi and others},
  journal={arXiv preprint arXiv:2504.02826},
  year={2025}
}

@article{wu2025kris,
  title={KRIS-Bench: Benchmarking Next-Level Intelligent Image Editing Models},
  author={Wu, Yongliang and Li, Zonghui and Hu, Xinting and Ye, Xinyu and Zeng, Xianfang and Yu, Gang and Zhu, Wenbo and Schiele, Bernt and Yang, Ming-Hsuan and Yang, Xu},
  journal={arXiv preprint arXiv:2505.16707},
  year={2025}
}

@inproceedings{luo2025mmmg,
  title={Mmmg: A massive, multidisciplinary, multi-tier generation benchmark for text-to-image reasoning},
  author={Luo, Yuxuan and Yuan, Yuhui and Chen, Junwen and Cai, Haonan and Yue, Ziyi and Yang, Yuwei and Daha, Fatima Zohra and Li, Ji and Lian, Zhouhui},
  booktitle={The Thirty-ninth Annual Conference on Neural Information Processing Systems Datasets and Benchmarks Track},
  year={2025}
}

@article{chang2025sridbench,
  title={SridBench: Benchmark of Scientific Research Illustration Drawing of Image Generation Model},
  author={Chang, Yifan and Feng, Yukang and Sun, Jianwen and Ai, Jiaxin and Li, Chuanhao and Zhou, S Kevin and Zhang, Kaipeng},
  journal={arXiv preprint arXiv:2505.22126},
  year={2025}
}

@article{wang2025genexam,
  title={GenExam: A Multidisciplinary Text-to-Image Exam},
  author={Wang, Zhaokai and Yin, Penghao and Zhao, Xiangyu and Tian, Changyao and Qiao, Yu and Wang, Wenhai and Dai, Jifeng and Luo, Gen},
  journal={arXiv preprint arXiv:2509.14232},
  year={2025}
}

@article{zhang2025large,
  title={Large multimodal models evaluation: a survey},
  author={Zhang, Zicheng and Wang, Junying and Wen, Farong and Guo, Yijin and Zhao, Xiangyu and Fang, Xinyu and Ding, Shengyuan and Jia, Ziheng and Xiao, Jiahao and Shen, Ye and others},
  journal={Science China Information Sciences},
  volume={68},
  number={12},
  pages={221301},
  year={2025},
  publisher={Springer}
}

@article{worldgenbench,
  title={WorldGenBench: A World-Knowledge-Integrated Benchmark for Reasoning-Driven Text-to-Image Generation},
  author={Zhang, Daoan and Jiang, Che and Xu, Ruoshi and Chen, Biaoxiang and Jin, Zijian and Lu, Yutian and Zhang, Jianguo and Yong, Liang and Luo, Jiebo and Luo, Shengda},
  journal={arXiv preprint arXiv:2505.01490},
  year={2025}
}

@article{Qwen3-VL,
      title={Qwen3-VL Technical Report}, 
      author={Shuai Bai and Yuxuan Cai and Ruizhe Chen and Keqin Chen and Xionghui Chen and Zesen Cheng and Lianghao Deng and Wei Ding and Chang Gao and Chunjiang Ge and Wenbin Ge and Zhifang Guo and Qidong Huang and Jie Huang and Fei Huang and Binyuan Hui and Shutong Jiang and Zhaohai Li and Mingsheng Li and Mei Li and Kaixin Li and Zicheng Lin and Junyang Lin and Xuejing Liu and Jiawei Liu and Chenglong Liu and Yang Liu and Dayiheng Liu and Shixuan Liu and Dunjie Lu and Ruilin Luo and Chenxu Lv and Rui Men and Lingchen Meng and Xuancheng Ren and Xingzhang Ren and Sibo Song and Yuchong Sun and Jun Tang and Jianhong Tu and Jianqiang Wan and Peng Wang and Pengfei Wang and Qiuyue Wang and Yuxuan Wang and Tianbao Xie and Yiheng Xu and Haiyang Xu and Jin Xu and Zhibo Yang and Mingkun Yang and Jianxin Yang and An Yang and Bowen Yu and Fei Zhang and Hang Zhang and Xi Zhang and Bo Zheng and Humen Zhong and Jingren Zhou and Fan Zhou and Jing Zhou and Yuanzhi Zhu and Ke Zhu},
	  journal={arXiv preprint arXiv:2511.21631},
      year={2025}
}

@misc{sun2025t2ireasonbenchbenchmarkingreasoninginformedtexttoimage,
      title={T2I-ReasonBench: Benchmarking Reasoning-Informed Text-to-Image Generation}, 
      author={Kaiyue Sun and Rongyao Fang and Chengqi Duan and Xian Liu and Xihui Liu},
      year={2025},
      eprint={2508.17472},
      archivePrefix={arXiv},
      primaryClass={cs.CV},
      url={https://arxiv.org/abs/2508.17472}, 
}

@article{ye2025imgedit,
  title={Imgedit: A unified image editing dataset and benchmark},
  author={Ye, Yang and He, Xianyi and Li, Zongjian and Lin, Bin and Yuan, Shenghai and Yan, Zhiyuan and Hou, Bohan and Yuan, Li},
  journal={arXiv preprint arXiv:2505.20275},
  year={2025}
}

@misc{hu2024ellaequipdiffusionmodels,
      title={ELLA: Equip Diffusion Models with LLM for Enhanced Semantic Alignment}, 
      author={Xiwei Hu and Rui Wang and Yixiao Fang and Bin Fu and Pei Cheng and Gang Yu},
      year={2024},
      eprint={2403.05135},
      archivePrefix={arXiv},
      primaryClass={cs.CV},
      url={https://arxiv.org/abs/2403.05135}, 
}

@misc{ghosh2023genevalobjectfocusedframeworkevaluating,
      title={GenEval: An Object-Focused Framework for Evaluating Text-to-Image Alignment}, 
      author={Dhruba Ghosh and Hanna Hajishirzi and Ludwig Schmidt},
      year={2023},
      eprint={2310.11513},
      archivePrefix={arXiv},
      primaryClass={cs.CV},
      url={https://arxiv.org/abs/2310.11513}, 
}

@misc{heusel2018ganstrainedtimescaleupdate,
      title={GANs Trained by a Two Time-Scale Update Rule Converge to a Local Nash Equilibrium}, 
      author={Martin Heusel and Hubert Ramsauer and Thomas Unterthiner and Bernhard Nessler and Sepp Hochreiter},
      year={2018},
      eprint={1706.08500},
      archivePrefix={arXiv},
      primaryClass={cs.LG},
      url={https://arxiv.org/abs/1706.08500}, 
}

@misc{salimans2016improvedtechniquestraininggans,
      title={Improved Techniques for Training GANs}, 
      author={Tim Salimans and Ian Goodfellow and Wojciech Zaremba and Vicki Cheung and Alec Radford and Xi Chen},
      year={2016},
      eprint={1606.03498},
      archivePrefix={arXiv},
      primaryClass={cs.LG},
      url={https://arxiv.org/abs/1606.03498}, 
}

@misc{yu2022scalingautoregressivemodelscontentrich,
      title={Scaling Autoregressive Models for Content-Rich Text-to-Image Generation}, 
      author={Jiahui Yu and Yuanzhong Xu and Jing Yu Koh and Thang Luong and Gunjan Baid and Zirui Wang and Vijay Vasudevan and Alexander Ku and Yinfei Yang and Burcu Karagol Ayan and Ben Hutchinson and Wei Han and Zarana Parekh and Xin Li and Han Zhang and Jason Baldridge and Yonghui Wu},
      year={2022},
      eprint={2206.10789},
      archivePrefix={arXiv},
      primaryClass={cs.CV},
      url={https://arxiv.org/abs/2206.10789}, 
}

@misc{huang2023smarteditexploringcomplexinstructionbased,
      title={SmartEdit: Exploring Complex Instruction-based Image Editing with Multimodal Large Language Models}, 
      author={Yuzhou Huang and Liangbin Xie and Xintao Wang and Ziyang Yuan and Xiaodong Cun and Yixiao Ge and Jiantao Zhou and Chao Dong and Rui Huang and Ruimao Zhang and Ying Shan},
      year={2023},
      eprint={2312.06739},
      archivePrefix={arXiv},
      primaryClass={cs.CV},
      url={https://arxiv.org/abs/2312.06739}, 
}

@misc{niu2025wiseworldknowledgeinformedsemantic,
      title={WISE: A World Knowledge-Informed Semantic Evaluation for Text-to-Image Generation}, 
      author={Yuwei Niu and Munan Ning and Mengren Zheng and Weiyang Jin and Bin Lin and Peng Jin and Jiaqi Liao and Chaoran Feng and Kunpeng Ning and Bin Zhu and Li Yuan},
      year={2025},
      eprint={2503.07265},
      archivePrefix={arXiv},
      primaryClass={cs.CV},
      url={https://arxiv.org/abs/2503.07265}, 
}

@misc{Gemini-3-Flash,
  title={Gemini 3 Flash: frontier intelligence built for speed},
  author={Google},
  howpublished={\url{https://blog.google/products-and-platforms/products/gemini/gemini-3-flash/}},
  year={2025}
}

@misc{Nano-Banana-Pro,
  title={Introducing Nano Banana Pro},
  author={Google},
  howpublished={\url{https://blog.google/innovation-and-ai/products/nano-banana-pro/}},
  year={2025}
}

@misc{Nano-Banana-2,
  title={Nano Banana 2: Combining Pro capabilities with lightning-fast speed},
  author={Google},
  howpublished={\url{https://blog.google/innovation-and-ai/technology/ai/nano-banana-2/}},
  year={2025}
}

@misc{Artificial-Analysis,
  title={Image Editing Leaderboard},
  author={Artificial Analysis},
  howpublished={\url{https://artificialanalysis.ai/image/leaderboard/editing/}},
  year={2026}
}

@article{GPT-Image-1,
  title={GPT-Image-1},
  author={OpenAI},
  journal={https://openai.com/index/image-generation-api/},
  year={2025}
}

@article{GPT-Image-1.5,
  title={GPT-Image-1.5},
  author={OpenAI},
  journal={
https://openai.com/zh-Hans-CN/index/new-chatgpt-images-is-here/},
  year={2025}
}

@misc{li2024surveymultimodalcompositeediting,
      title={A Survey of Multimodal Composite Editing and Retrieval}, 
      author={Suyan Li and Fuxiang Huang and Lei Zhang},
      year={2024},
      eprint={2409.05405},
      archivePrefix={arXiv},
      primaryClass={cs.CV},
      url={https://arxiv.org/abs/2409.05405}, 
}

\end{document}